%% file: fairness.tex
\documentclass{article} 
\usepackage{iclr2020_conference,times}

\input{math_commands.tex}

\usepackage{multirow}
\usepackage{graphicx}
\usepackage[section]{placeins}
\usepackage{hyperref}
\usepackage{url}

\usepackage{array}

\title{Towards classification parity across cohorts}

\author{Aarsh Patel \\
Department of Computer Science\\
University of Massachusetts\\
Pittsburgh, PA 15213, USA \\
\texttt{aarshpatel@umass.edu} \\
\And
Rahul Gupta \\
Amazon Alexa \\
Cambridge, Massachusetts\\
\texttt{gupra@amazon.com} \\
\And
Mukund Harakere \\
Amazon Alexa \\
Cambridge, Massachusetts\\
\texttt{harakere@amazon.com} \\
\And
Satyapriya Krishna  \\
Amazon Alexa \\
Cambridge, Massachusetts\\
\texttt{satyapk@amazon.com} \\
\And
Aman Alok \\
Amazon Alexa \\
Cambridge, Massachusetts\\
\texttt{alokaman@amazon.com} \\
\And
Peng Liu \\
Amazon Alexa \\
Cambridge, Massachusetts\\
\texttt{liupng@amazon.com} \\
}

\iclrfinalcopy 
\begin{document}

\maketitle

\begin{abstract}
Recently, there has been a lot of interest in ensuring algorithmic fairness in machine learning where the central question is how to prevent sensitive information (e.g. knowledge about the ethnic group of an individual) from adding ‘unfair’ bias to a learning algorithm (\cite{feldman2015certifying}, \cite{zemel2013learning}). This has led to several debiasing algorithms on word embeddings (\cite{qian2019reducing} , \cite{bolukbasi2016man}), coreference resolution (\cite{zhao2018gender}), semantic role labeling (\cite{zhao2017men}), etc. Most of these existing work deals with explicit sensitive features such as gender, occupations or race which doesn't work with data where such features are not captured due to privacy concerns. In this research work, we aim to achieve classification parity across explicit as well as implicit sensitive features. We define explicit cohorts as groups of people based on explicit sensitive attributes provided in the data (age, gender, race) whereas implicit cohorts are defined as groups of people with similar language usage. We obtain implicit cohorts by clustering embeddings of each individual trained on the language generated by them using a language model. We achieve two primary objectives in this work :  [1.] We experimented  and discovered classification performance differences across cohorts based on implicit and explicit features , [2] We improved classification parity by introducing modification to the loss function aimed to minimize the range of model performances across cohorts.



\end{abstract}

\input{introduction}
\input{proposed_solutions}

\input{experimental_results}
\input{conclusions}

\bibliography{iclr2020_conference}
\bibliographystyle{iclr2020_conference}

\appendix

\section{Fairness Definitions}
\label{appen:A}
\begin{itemize}
	\item \textbf{Demographic Parity }: It ensures that the likelihood of an outcome is independent of whether the person belongs to any social group, known as sensitive attribute, for instance, gender is a sensitive attribute. (\cite{verma2018fairness})
	\item \textbf{Equalized Odds } : It requires to have the same rate for true positive as wells as false positives for all groups from sensitive attribute (eg. male and female). (\cite{hardt2016equality})
\end{itemize}

\section{Learning Implicit Cohorts}
\label{appen:B}
We learnt implicit cohorts by learning embeddings for individuals based on their historical utterances that are representative of their language usage and vocabulary usage. Each individual was considered as another token and appended at the beginning of the sentence which was fed to an LSTM based language model \cite {sundermeyer2012lstm} to train. These trained embeddings were then clustered using k-means clustering to extract cohorts.

\section{Performance across cohorts}
\label{appen:C}
\begin{figure}[h]
	\caption{ 	\label{fig:yd} Model performances across cohorts in Yelp Dataset}
	\centering
	\includegraphics[width=0.8\textwidth]{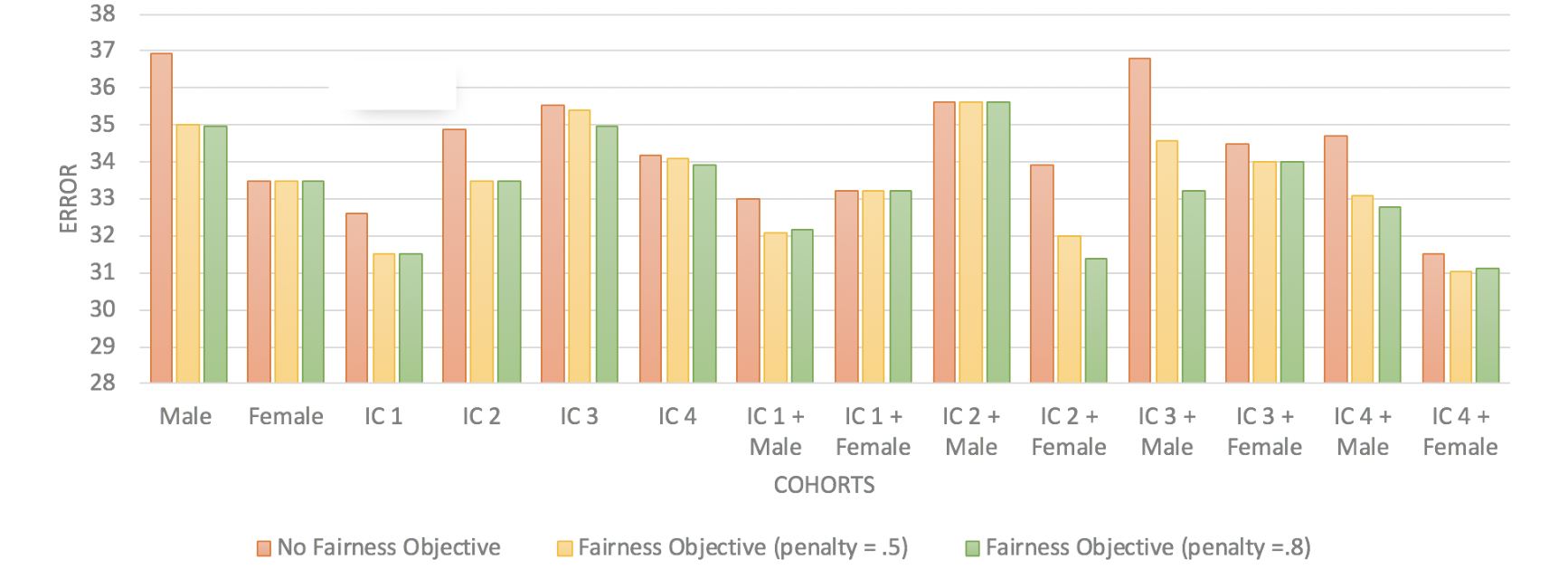}
\end{figure}

\begin{figure}[h]
	\caption{	\label{fig:id} Model performances across cohorts in Internal Dataset}
	\centering
	\includegraphics[width=1.1\textwidth]{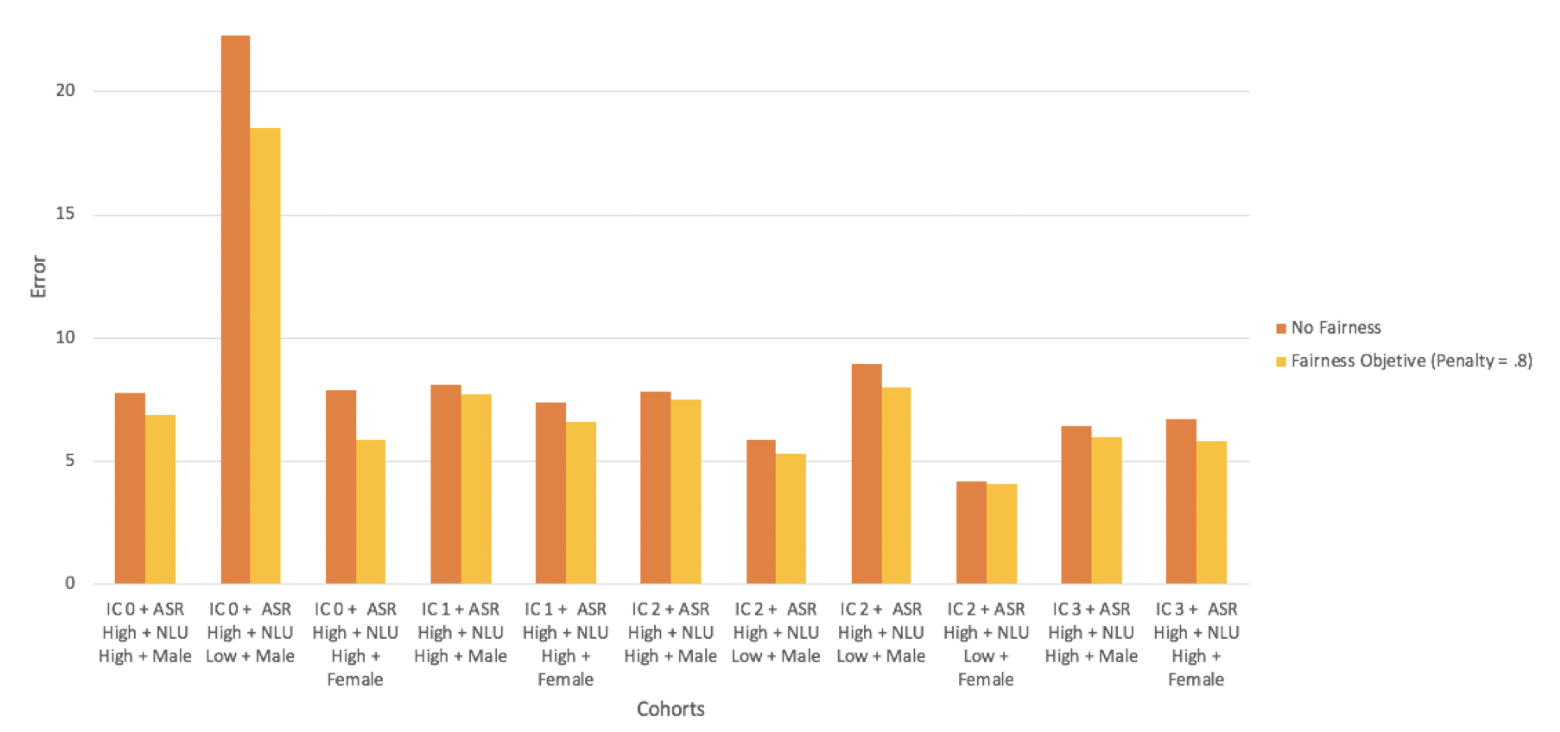}
	
\end{figure}

\begin{figure}[h]
	\caption{	\label{fig:tpd} Model performances across cohorts in TrustPilot Dataset}
	\centering
	\includegraphics[width=1.1\textwidth]{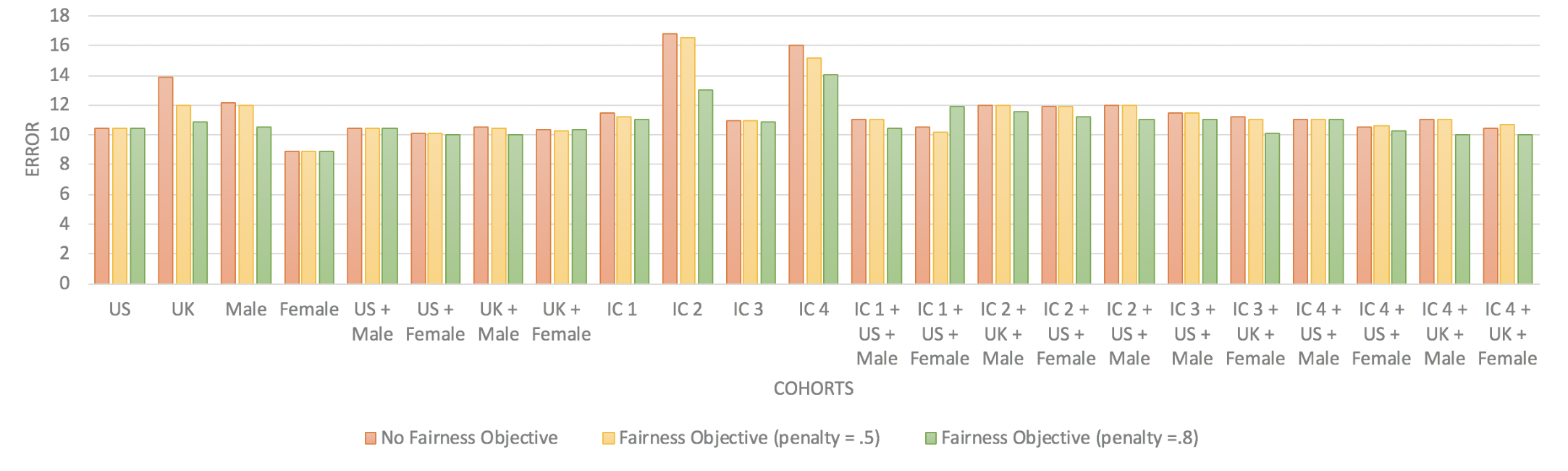}
\end{figure}

We look into the model performances across cohorts after adding the penalty term in the loss function. In order to check the consistency of our model, we created a spectrum of cohorts by randomly combining explicit and implicit cohorts which we had extracted before. For instance, we combined implicit cohort 1 (IC 1) and gender to create another cohort. The model performances over all the cohorts are shown in figures \ref {fig:yd} \ref {fig:id} \ref {fig:tpd} . We observe the parity improve for almost all the cohorts as we increase $\lambda$, although we also note the decline in model performance for those cohorts which is a typical behavior after adding fairness constraints. 

\end{document}

%% file: math_commands.tex
\usepackage{amsmath,amsfonts,bm}

\def\eqref#1{equation~\ref{#1}}

\def\1{\bm{1}}

\DeclareMathAlphabet{\mathsfit}{\encodingdefault}{\sfdefault}{m}{sl}
\SetMathAlphabet{\mathsfit}{bold}{\encodingdefault}{\sfdefault}{bx}{n}



%% file: introduction.tex
\FloatBarrier
\section{Introduction}

Machine learning has proven to be useful in several applications such as speech recognition (\cite{graves2013speech}),  image classification (\cite{krizhevsky2012imagenet}), reading comprehension (\cite{hermann2015teaching}), etc and growing in other critical application such as healthcare (\cite{esteva2019guide}). With such rapid increase in data driven solutions making important decisions in our lives, it is worth discussing the possibility of bias in these models. Apart from the  generally discussed algorithmic bias, there are several other types of bias such as, representational bias (\cite{suresh2019framework}) due to the lack of diversity in data samples, historical bias  (\cite{suresh2019framework}) caused due to the existing societal biases unknowingly making its way to the data generation process, etc that can negatively impact decision making. 

Recently, there has been a lot of work in removing biases through debiasing techniques motivated by algorithmic fairness (\cite{gonen2019lipstick} \cite{zhao2018learning}). All these models works only on explicit attributes such as gender or occupation which comes with two downsides. First, explicit attributes may not be available in a generic dataset due to privacy concerns. Secondly, gender might not be the only attribute causing unfair bias in the model. Hence, we experimented with explicit (gender) as well as implicit (hidden) sensitive attributes and improved classification parity for both of them.


\FloatBarrier

%% file: proposed_solutions.tex
\section{Proposed Method}

\subsection{Background }
\label{background}
 In the last few years, several fairness definitions (\cite {hardt2016equality}\cite{dwork2012fairness}\cite{kusner2017counterfactual}\cite{berk2018fairness}) have been proposed taking different viewpoints under consideration. In this research work, we worked only on group level fairness definitions for two reasons. First, we are aiming to provide classification parity across cohorts before going down to individual performance parity. Second, its impossible to satisfy all the fairness definitions at once (\cite{kleinberg2016inherent}) without complex constraints, hence, we experimented with group level fairness for simplicity, i.e, demographic parity (\cite{verma2018fairness}) and equalized odds(\cite{hardt2016equality}).  There are multiple ways of implementing these definitions to debias our model and reach classification parity. In our work, we try to implement them by making modification to the loss function only and avoid any complex model architecture changes. This helps in scaling our solution to different applications.

\subsection{Objective }

With the goal to achieve classification parity across multiple cohorts, we propose the optimization of loss function in equation~\ref{eq:loss}, where
$(\bm x,y)$ represents the pair of features and labels, respectively, in a dataset $\mathcal D$.  $l(y, f(\bm x))$ is the loss incurred by a trainable function $f(\bm x)$. $\mathcal D_i$ represents the cohort $i$ in the dataset and $\bm x_i, y_i$ are the datapoints belonging to the cohort $i$. 
The loss function has two components as depicted in the equation: the first component is an overall loss over the entire dataset and the second component enforces parity across cohorts.
$\lambda$ is the weightage given to parity component when compared to the loss over the entire dataset.

\begin{equation}\label{eq:loss}
\mathcal L = \sum_{(\bm x,y) \in \mathcal D} l(y, f(\bm x)) + \lambda  max_{i,j} \Big| \sum_{(\bm x_i, y_i) \in \mathcal D_i} l(y_i, f(\bm x_i)) -  \sum_{(\bm x_j, y_j) \in \mathcal D_j} l(y_j, f(\bm x_j)) \Big|  
\end{equation}

We chose to minimize the difference between best and worst performing cohorts for the sake of simplicity. We also considered an alternate strategy to evaluate every pair during optimization. However, this strategy is expensive and determining weights for each pairwise difference is non-trivial. 
Additionally, this loss formulation allows the cohorts $i,j$ used in the optimization to either be coarse level cohorts (e.g. gender) or fine grained cohorts (e.g. a specific gender belonging to a specific ethnic background and a specific country).
We use this inherent flexibility of the loss function to dynamically modify the granularity of cohorts during optimization.
We use stochastic gradient descent to minimize $\mathcal L$, and we note that during each iteration, one may obtain a different cohort pair $i,j$ in the parity loss component.


\subsection{Cohort definition}
\label{obj}
In the datasets of our interest, we use two broad categories of cohorts - explicit and implicit.
We provide a brief description of these below. 

\subsubsection{Explicit cohorts}
We define cohorts obtained based on attributes such as gender and location as explicit cohorts.
These attributes could be directly observed for each individual in the dataset. 

\subsubsection{Implicit cohorts}
We realize that available explicit attributes may not capture all the characteristics of a given individual.
For instance, individuals may have not reported their ethnic background and/or some attributes may have not been recorded during data collection.
To address this, we also determine implicit cohorts based on an individual's language usage.
We obtain vector representations for each individual based on the language generated by them using a language model.
In summary, we add an individual-id as additional token during language modeling task and use the {\it word-embedding} obtained for the individual-id token as their vector representation.
These vector representations are clustered to define implicit cohorts.

%% file: experimental_results.tex
\section{Experiment}

Our experiment followed a simple framework where we started with performance comparison across cohorts (implicit and explicit) and then retrained models with our proposed changes to the loss function. We measured classification parity in terms of the standard deviation of performance across cohorts.
\subsection{Datasets}
We used three datasets for our experiments and their details are listed below. 
\begin{center}
	\begin{table}[ht]
	\begin{tabular}{ || m{7em} | m{8cm} | m{3cm} || } 
		\hline
		Dataset & Description & Classification Task \\ [0.5ex] 
		\hline\hline
		\textbf{Yelp Dataset (YD) } \cite{yelp} & Contains reviews from different businesses with a 5 star rating attached to every review. We added gender component in this dataset by using a publicly available tool (\cite{genderize}) which predicts gender by the name of the individual &  Review rating prediction task  \\ [0.5ex]
		\hline
		\textbf{TrustPilot Dataset (TPD)}  \cite {hovy2015user} &  Contains reviews from different business across the US and UK with a 5 star rating along with user metadata such as gender and location. & Review rating prediction task \\ [0.5ex]
		\hline
		\textbf{ Internal Dataset (ID)} & Contains user utterances with labeled domain information (23 domain categories) along with metadata such as the gender, NLU(Natural Language Understanding) Score and ASR (Automatic Speech Recognition) score, which are the confidence scores of speech recognition and language understanding modules of the system & Domain Classification \\
			\hline
	 \end{tabular}
 \caption{\label{Tab:Tcr1} Data description with corresponding classification task }
 \end{table}
\end{center}
\subsection {Disparity in Model Performance}

In the first segment of our experiments, we computed model performance across different cohorts for the corresponding tasks mentioned in Table \ref{Tab:Tcr1} for each dataset.

\subsubsection{Explicit Cohorts}
Explicit cohorts are based on explicit sensitive attributes such as gender or location. In the case of internal dataset, we used another set of explicit attributes, i.e, ASR (Automatic Speech Recognition) and NLU (Natural Language Understanding) scores since it represents the nativity of cohorts, i.e, people speaking a specific dialect will have similar ASR scores. We create two cohorts for each of these scores based on a predefined threshold($t$) and called it "NLU High" cohort with samples having NLU scores higher that $t$ and "NLU Low" cohort otherwise, similarly for ASR score. We observed disparities across all of the explicit cohorts and listed in table \ref{Tab:Tcra}

\begin{center}
		\begin{table}[ht]
	\begin{tabular}{ || m{5em} | m{3cm}  | m{3cm} | m{2cm} || } 
		\hline
		Dataset & Sensitive Attribute &  & Accuracy \\ [0.5ex] 
		\hline\hline
		 \textbf{YD} & 	
		                                     \multirow{2}{*} {Gender} 
		                                     &  Male  &  0.6306  \\
			\cline{3-4}																						
										& 											 & Female & 0.665 \\
		\hline
		\multirow{2}{*}{\textbf{TPD}} & 	\multirow{2}{*} {Location (country)}  & UK Cohort  &  0.8612  \\
																																		\cline{3-4}
																																	&	& US Cohort & 0.8955  \\ 
																																		\cline{2-4}
																			& 	\multirow{2}{*} {Gender} 															
																																		& Male & 0.8787  \\ 
																																		\cline{3-4}
																																&		& Female & 0.9109  \\ 
		\hline 
		\multirow{6}{*}{\textbf{ID}} & 	\multirow{2}{*} {Gender} & Male  &    \\
																															\cline{3-4}
																														&	& Female & 0.038  \\ 
																															\cline{2-4}
																				& 	\multirow{2}{*} {ASR Score} 											
																															& ASR High &   \\ 
																															\cline{3-4}
																													&		& ASR Low & 0.049  \\ 
																															\cline{2-4}
																				& 	\multirow{2}{*} {NLU Score}											
																															& NLU High &  \\ 
																															\cline{3-4}
																													&		& NLU Low & 0.0514  \\ 
		\hline 
	\end{tabular}
\caption{	\label{Tab:Tcra}  Model performance over explicit cohorts. We  report performance differences for internal dataset (ID) only. For instance, abs(Accuracy(Male Cohort) - Accuracy(Female Cohort)) = 0.038 }
 \end{table}
\end{center}
\subsubsection{Implicit Cohorts}
Implicit cohorts are based on implicit sensitive features which could be in terms of the linguistic differences based on regions (\cite{labov1997national}). We extracted such implicit cohort by training a language model for individuals based on their historical utterances, more details in Appendix \ref{appen:B}. We used k-means to cluster the embeddings, used k = 4 for the experiments. The performance difference across the cohorts are listed in table \ref{Tab:Tcr1a}.

  \begin{center}
  \begin{table}[ht]
  \begin{tabular}{ || m{5em} | m{3.1cm}  | m{3.1cm} | m{2cm} | m{2cm} || } 
  	\hline
  	Dataset & Implicit Cohort-1  (IC 1) & Implicit Cohort-2 (IC 2)  & Implicit Cohort-3 (IC 3) & Implicit Cohort-4 (IC 4)  \\ [0.5ex] 
  	\hline\hline
  	\textbf{YD} & 	0.674 &  0.6511  &  0.6445 & 0.658  \\
  	
  	\hline 
  	\textbf{TPD} & 	0.885 &  0.832  &  0.8901 & 0.8402  \\
  	
  	\hline
  		\textbf{ID} & 	0.0205 &    &   &   \\ 	
  	\hline 

  \end{tabular}
	 
	 \caption{	\label{Tab:Tcr1a}  Model performance of implicit cohorts. We report std. deviation of accuracy on domain classification task for ID only.   }
\end{table}
  \end{center}

\subsection{ Classification Parity  }
In order to reduce disparities discovered in the last section, we retrained our model with the proposed changes to the loss function and measured the disparity in terms of the standard deviation of model performances across cohorts over test set. We experimented with different values of $\lambda $ in equation \ref{eq:loss} to observe the changes it causes to the parity and the overall model performance. We computed the standard deviation over a spectrum of cohorts by grouping multiple explicit or implicit sensitive attributes which gives a better picture about the robustness of our solution, more details in Appendix \ref{appen:C}.

  \begin{center}
	\begin{table}[ht]
		
		\begin{tabular}{ || m{5em} | m{3cm}  | m{3cm} | m{2cm} || } 
			\hline
			Dataset & No penalty  & $\lambda$ = 0.5  & $\lambda$ = 0.8   \\ [0.5ex] 
			\hline\hline
			\textbf{ID} & 	2.96 &  2.8  &  2.62 \\
			
			\hline 
			\textbf{TPD} & 	1.81 &  1.61  &  1.06  \\
			
			\hline
			\textbf{YD} & 	1.55 &  1.42  &  1.38   \\ 	
			\hline 
			
		\end{tabular}
		\caption{	\label{Tab:Tcr2b}  Standard deviation of model performance across cohorts for $\lambda$ = 0, 0.5, 0.8 }
	\end{table}
\end{center}
As shown in the table \ref{Tab:Tcr2b}, we observe a drop in the standard deviation by 10-12 \% with increased value of $\lambda $.



%% file: conclusions.tex
\section{Conclusion}

In this work, we experimented and found disparities in model performance across explicit and implicit cohorts. In order to fix these disparities, we added a penalty term in the loss function with the goal to minimize performance difference between the best and worst performing cohort.  Our experiments showed that this change reduces disparity across explicit and implicit cohorts. We also see improvements in parity as we increase the weight for this penalty term with minimal impact to the overall model performance for multiple cohorts.  